# A Hybrid Approach Based Segmentation Technique for Brain Tumor in MRI Images

D. Anithadevi[1] and K. Perumal[2]

[1]Research scholar, Department of Computer Applications, Madurai Kamaraj University, Madurai.
danithatce@gmail.com

[2]Associate Professor, Department of Computer Applications, Madurai Kamaraj University, Madurai.
perumalmkucs@gmail.com

## ABSTRACT

*Automatic image segmentation becomes very crucial for tumor detection in medical image processing. Manual and semi automatic segmentation techniques require more time and knowledge. However these drawbacks had overcome by automatic segmentation still there needs to develop more appropriate techniques for medical image segmentation. Therefore, we proposed hybrid approach based image segmentation using the combined features of region growing and threshold segmentation technique. It is followed by pre-processing stage to provide an accurate brain tumor extraction by the help of Magnetic Resonance Imaging (MRI). If the tumor has holes in it, the region growing segmentation algorithm can't reveal but the proposed hybrid segmentation technique can be achieved and the result as well improved. Hence the result used to made assessment with the various performance measures as DICE, Jaccard similarity, accuracy, sensitivity and specificity. These similarity measures have been extensively used for evaluation with the ground truth of each processed image and its results are compared and analyzed.*

## 1. INTRODUCTION

Hybrid Segmentation technique integrating two or more techniques which is efficiently giving better results than the segmentation algorithms working alone. This is all possible in the field of Image Processing, predominantly in the area of medical image segmentation [1, 2, 6 and 15]. Image segmentation means separating the objects from the background. Image segmentation acts as a heart to the classification technique. The, proposed system mainly focused on medical imaging to extract tumor and especially in MRI images. It has high-resolution and accurate positioning of soft and hard tissues, and is especially suitable for the diagnosis of brain tumors [1, 2 and 6]. So this type of imaging is more suitable to identify the brain lesions or tumor. Brain tumor is abnormal white tissues which can be differed from normal tissues. This could be identified by the structure of tissues. The MRI images which contains tumor that are shown below:

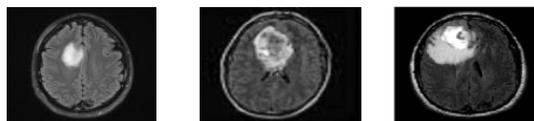

Figure 1: Representation of various tumor images.

Generally, tumors are looking like solid white tissues or consist of holes. Figure 1, seems the tumor variation and (a) represents a solid tumor, (b) the variations among tumor cells and (c) consist hole in the tumor part. Figure 1b, c are unable to identify by the region growing segmentation. Therefore, the threshold segmentation is intently combined with the region growing for result improvement.

## 2. RELATED WORKS

Gurjeet kaur Seerha et.al [3], worked with various segmentation algorithms. Now-a-days plenty of segmentation techniques are used for segment the images but there is not even a single method worked for all types of images. So they are discussed to develop a segmentation technique which is used to solve a single problem in all types of images. Kalaivani et.al [5], discussed about automatic seed selection for a region growing segmentation. In region growing segmentation seed point selection plays crucial role. The fully worked for a color image segmentation. And they worked on its algorithm's speed, noise immunity, atomicity, accuracy, etc. for a brain tumor image. Iraky khalifa, et.al [6], worked with hybrid method for the segmentation of MRI brain tumor images and they analyze its quality. Manoj kumarV et.al [9], has been discussed various segmentation algorithms for an MRI images and analyze the performance of those algorithms. V. Murali, et.al [10], discussed some segmentation algorithms and compares their performance with an MRI brain tumor images. N. A. Mat-Isa, et.al [11, 16], had worked on brain tumor MRI images [11] worked with seeded region growing algorithm[7, 14] and extracting features for classification from an image using segmentation. The features are useful for classification. H.P. Narkhede, et.al [4, 10, 12, 14, 22, 23], had review on various segmentation techniques and evaluates the performance of these techniques and analyzed its quality. Partha Sarathi Giri [13], had been worked with an digital image and extracting text information from a digital image. Rajesh Dass, et.al [15], worked on digital images with various segmentation algorithms. T.Romen Singh, et.al [17, 18], worked in threshold techniques; in [17] they especially worked in adaptive threshold segmentation. K.Srinivas, et.al [19], presented fuzzy measure with c means using automatic histogram threshold. As far as the segmentation, threshold value is essential for binarization, they calculated the value from histogram to apply this value in segmentation algorithm. Sukhjinder Singh, et.al [20], discussed to work with a matlab and to develop an image processing applications in matlab.

## 3. PROPOSED WORK

MRI brain image explicitly contains tumor portion is taken as an input image. This work contains three phases.

**Phase 1:** The pre-processing steps (i.e. Gray scale conversion, contrast enhancement) have done.
**Phase 2:** The results of single seed region growing and threshold segmentation results are multiplied.
**Phase 3:** Performance of the proposed system can be measured by the various quality metrics.

In region growing segmentation seed point selection is necessary step. Two types of seed point selections are in the seeded region growing.

(i) Semi automated seed point selection.
(ii) Fully automated seed point selection.

If seed point has selected manually during run time that called as semi-automatic segmentation. In fully automated segmentation, automated seed point selection is made. Automated seed point is selected in many ways as, centre pixel/fixed, random and high intensity seed points. In threshold segmentation, threshold value is more essential. It has two types as

   (i) Single value threshold
   (ii) Multi threshold

If the threshold segmentation works with the single threshold value which is called single value threshold segmentation otherwise, it's multi value threshold segmentation.

In this proposed system centre pixel/fixed seed point is utilized for region growing segmentation and single threshold value for threshold segmentation. By experimental and performance of this hybrid segmentation is measured which will perform well and gives expected output that is considered as a good segmentation to extract tumor as it is.

## 4. METHODOLOGY

In this system some useful methods are applied to achieve an expected result. These methods are ordered according their usage. The overall process flow of the proposed system is shown below:

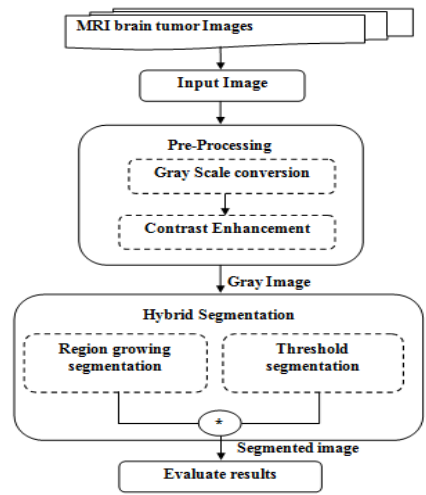

Figure 2: Flow Diagram of overall process.

### 4.1. Pre-Processing

The input images are abnormal brain tumor MRI images. Even the MRI image having high definition in visualizing soft tissues, there is a need of contrast enhancement [1]. Before this, gray scale conversion is needed because it reduces complexity than color image. Visually, in the image no noises can be identified. Some techniques have to be applied in the image those are median filter and stationary wavelet technique. These are helps to extract features from an image. Practically, the MRI images are enhanced using the combinations of contrast, median filter and stationary wavelet methods gave better result for segmentation.

### 4.2. Hybrid Segmentation Approach

Hybrid segmentation is a prime method and an appealing approach to attaining the targeted brain tumor image segmentation. The major problem in single seed region growing is inability to expose the holes in the tumor. So this problem is revealed by this hybrid segmentation technique. The expected result can be done by the combined techniques of single seed region growing and single threshold segmentation. The performance of this proposed hybrid technique result depends on similarity measures used by the method and its implementation. Before proceeding to the performance evaluation part, a brief methodology of the hybrid segmentation algorithms is given below:

### 4.2.1. Seed Region Growing

This technique is accomplished by using single seed point [14]. A single seed point or pixel is taken and all the neighboring pixels are related to this seed forms the region r. The following criteria involved in the region growing segmentation:

1. Seed point should be selected automatically.
2. Seed point must have minimum distance with its neighbor pixels.
3. For a predictable region, at least one seed should be generated to produce the region.
4. Seeds of different regions should be disconnected.

During implementation, the minimum pixel distance is taken as a default process. The region is iteratively grown by evaluating all non-distributed neighboring pixels to the region. The variation between the value of pixel intensity and the mean of region is used as a measure of similarity. The pixel with the minimum difference measured is allocated to the particular region. This process stops when the intensity variation between region mean and new region become larger than a certain seed. Finally the output image is given by combining both the regions. Thus the segmented image using single seed region growing is formed as a binary image.

From the process which are less working favorable with the region growing segmentation so, to improve its performance by using threshold segmentation. Finally performance is analyzed by the appropriate measures.

### 4.2.2. Threshold segmentation method

Threshold is one of the image segmentation methods which are predominantly used to selecting an appropriate threshold value T, the gray image can be converted into binary image. The outcome should contain the entire decisive information about the objects position and shape. The threshold value (T) is obtained from the gray image and it can be classified into black (0), and White (1). The global threshold is used and the objective function is

$$g(x, y) = \begin{cases} 1 \ if \ f(x,y) > T \\ 0 \ if \ f(x,y) \leq T \end{cases}$$

Where $f(x, y)$ is an input image, g(x,y) threshold/segmented image, T threshold value.

The process of threshold segmentation is:
1. Initial estimate of threshold T.
2. Perform segmentation using T
    (i) $P_1$, pixels brighter than T
    (ii) $P_2$, pixels darker than T.
3. Apply average intensities $m_1$ and $m_2$ of $P_1$ and $P_2$.
4. Compute new threshold value

$$T_{new} = \frac{m_1 + m_2}{2}$$

5. If │T - T$_{new}$│ >ΔT, repeat step 2.
   Otherwise stop the process.

where $m_1$ and $m_2$ are mean of intensities, $P_1$ and $P_2$ is a probability of brighter and darker pixels and T and T$_{new}$ are the thresholds.

While it is natural and convergence, it can get enclosed in local mean and results can differ considerably depending on the thresholds. It can be obtaining the significant result when its combined with region growing segmentation.

### 4.2.3. Performance measures for analysis

The common performance measures of segmentations are: Jaccard distance, dice measures, Accuracy, Sensitivity, Specificity, Precision, Recall, F-Measure and G-mean of ground truth images and segmented images.

**a. Jaccard distance**

The Jaccard distance can be measured dissimilarity between expected and observed images, is complementary to the Jaccard coefficient and is acquired by subtracting the Jaccard coefficient from one. If $x = (x_1, x_2, \ldots x_n)$ and $y = (y_1, y_2, \ldots y_n)$ are two vectors with all real$(x_i, y_i) \geq 0$, then their Jaccard similarity coefficient is defined as

$$J(x, y) = \frac{\sum_i \min(x_i, y_i)}{\sum_i \max(x_i, y_i)}$$

Jaccard distance

$$d_J(x, y) = 1 - J(x, y)$$

If *X* and *Y* are both empty, we define *J(X, Y)* = 1. $0 \leq J(x, y) \leq 1$. Jaccard coefficient is used to measure similarity [16]. It computes the similarity between segmented and ground truth image. Jaccard's distance values lies between 0 and 1. Jaccard distance reaches its best value at 1 and worst value at 0. The objects having a 0 value along with their variables show a lower similarity.

**b. Dice Coefficient**

The dice coefficient D is one of a number of measures of the extent of spatial overlap between two binary images. It is commonly used in performance measures of segmentation and gives more weighting to instances where the two images agree [16]. Its values range between 0 (no overlap) and 1 (expected image).

$$D = \frac{2(AG)}{(AG + AG)} \times 100.$$

**c. Accuracy, Precision and Recall**

The **accuracy** regards to systematic errors and **precision** is related to random errors. The accuracy is directly proposed to true results consider both true positives and true negatives among the total number of cases scrutinized. To make the context clear by the semantics, it is frequently defined as the "rand accuracy". It is a parameter of the test.

$$\text{Accuracy} = \frac{number\ of\ true\ positives + number\ of\ true\ negatives}{number\ of\ true\ positives + false\ positives + false\ negatives + true\ negatives}$$

Conversely, precision (i.e. positive predictive value) is referred as the direct proportion of the true positives against all the results of positives (both true positives and false positives).

$$\text{Precision} = \frac{number\ of\ true\ positives}{number\ of\ true\ positives + false\ positives}$$

The high **precision** means that an algorithm returned substantially more appropriate results than irrelevant, whereas high **recall** means that an algorithm returned most of the pertinent results.

- TP (True Positive):
  It denotes the test result is one that detects the condition when condition is present.

- FP (False Positive):
  It signifies the test result that doesn't detect the condition when the condition is absent.

- FN (False Negative):
  It represents the test result that detects the condition when the condition is absent.

- TN (True Negative):
  It refers to the test result that doesn't detect the condition when condition is present.

Table 1: Test Statistics.

|  |  | Condition (ground truth Image) | |
|---|---|---|---|
|  |  | Present (Tumor) | Absent (non-tumor) |
| Test (Observed Images) | Tumor | True Positive | False Positive |
|  | Non-Tumor | False Negative | True Negative |

**d. F Measure**

In statistical analysis of binary segmentation, the $F_1$ score (also F-score or F-measure) is a measure of a test's accuracy. The $F_1$ score can be construed as a weighted average of the precision and recall, whereas an $F_1$ score reaches its best value at 1 or nearer to 1 and worst value at 0 or nearer to 0. It can be defined as

$$F_1 = 2 \cdot \frac{precision \cdot recall}{precision + recall}$$

**e. G-measure**

While the F-measure is the harmonic mean of Recall and Precision, the G-measure is the geometric mean.

$$G = \sqrt{precision \cdot recall}$$

## 5. RESULT AND ANALYSIS

### 5.1. Results

The hybrid segmentation technique is used to segment the MRI brain tumor images. An image is first pre-processing by converting into gray image and it is denoised using various techniques such as median filter and Stationary wavelet transforms. Already we worked with these combine features of denoising technique [2] which gave the results more favourable for this segmentation. This pre-processed image is segmented using the combined features of single seed region growing and threshold segmentation methods. The obtained results are analyzed by the quality measures.

## 5.2. Analysis

The result of hybrid segmentation technique is compared with the combinations of region growing and threshold segmentation techniques. In this analysis, similarity of the ground truth image and various observed segmented images are evaluated. In figure 3, $1^{st}$ row images are the output of region growing segmentation, $2^{nd}$ row represents the result of threshold segmentation, and $3^{rd}$ row shows the result of hybrid segmentation.

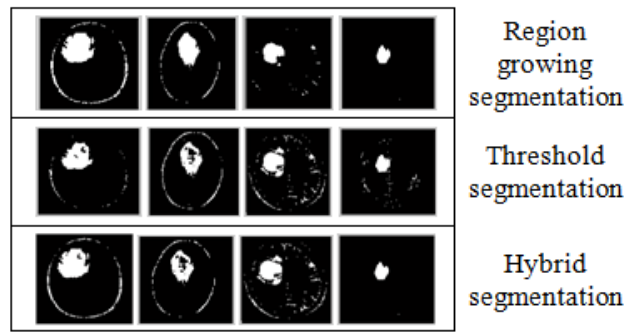

Figure 3: Segmentation results.

Figure 3, clearly shows the new hybrid segmentation method can able to generate the expected results by experimentally.

Image quality or performances are measured by the Jaccard distance and dice similarity. These measures purely designated to find the similarity between two binary images. Both Dice and Jaccard distance must higher value to facilitate the greater similarity between ground-truth segmented image and observed image. Therefore, both the values should be nearer 1, denotes the higher similarity.

Table 2: Jaccard and Dice for various segmentation techniques.

| Algorithms | Images | Jaccard distance | Dice |
|---|---|---|---|
| Region growing | Image 1 | 0.7654 | 0.6745 |
|  | Image 2 | 0.7809 | 0.7342 |
|  | Image 3 | 0.8678 | 0.8654 |
|  | Image 4 | 0.8234 | 0.8000 |
| Threshold | Image 1 | 0.8000 | 0.7476 |
|  | Image 2 | 0.9078 | 0.8405 |
|  | Image 3 | 0.7250 | 0.7125 |
|  | Image 4 | 0.8527 | 0.8800 |
| Hybrid | Image 1 | 0.8543 | 0.9954 |
|  | Image 2 | 0.9152 | 0.9002 |
|  | Image 3 | 0.9197 | 0.8923 |
|  | Image 4 | 0.9572 | 0.9707 |

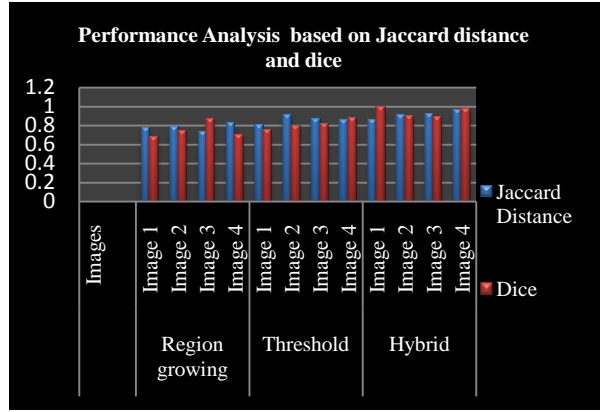

Figure 4: Performance analysis of various segmentations.

- In table 2, Jaccard distance and dice should be high similarity and similarity must be high for better segmentation.
- Jaccard distance and dice having the range 0 to 1.
- These measurements measure the percentage of correct decisions made by the algorithms and the maximum value indicates the maximum similarity.
- As per the results of Figure 4 & table 2, hybrid segmentation proves to be efficient by showing high Jaccard distance and Dice values than all other Methods.

Table 3: Performance analysis based on various segmentation techniques.

| Methods | Images | Accuracy | Precision | Recall | F-measure | G-measure |
|---|---|---|---|---|---|---|
| Region Growing | Image 1 | 0.7342 | 0.6745 | 0.7112 | 0.6996 | 0.8345 |
| | Image 2 | 0.6978 | 0.7342 | 0.7654 | 0.7908 | 0.7890 |
| | Image 3 | 0.7779 | 0.8654 | 0.8543 | 0.7809 | 0.8100 |
| | Image 4 | 0.8090 | 0.7000 | 0.8117 | 0.8099 | 0.8587 |
| Threshold | Image 1 | 0.7250 | 0.7476 | 0.8990 | 0.8000 | 0.9180 |
| | Image 2 | 0.8527 | 0.7905 | 0.9078 | 0.7090 | 0.8999 |
| | Image 3 | 0.8123 | 0.8125 | 0.7888 | 0.8609 | 0.8678 |
| | Image 4 | 0.8234 | 0.8800 | 0.8101 | 0.9000 | 0.8911 |
| Hybrid | Image 1 | 0.9249 | 0.9777 | 0.9034 | 0.9927 | 0.8769 |
| | Image 2 | 0.9781 | 0.9342 | 0.9456 | 0.9314 | 0.9235 |
| | Image 3 | 0.9502 | 0.9523 | 0.8210 | 0.8999 | 0.9156 |
| | Image 4 | 0.9827 | 0.9860 | 0.9987 | 0.9797 | 0.9843 |

As far as this table 3 concerned the performance results, hybrid segmentation algorithm has higher accuracy, precision, recall, f and g measure values. While considering these measures nearer to 1 is efficient. Especially the accuracy should be nearer or equal to 1 that shows the efficiency of the hybrid method. But on taking both precision and accuracy measures into accounts the hybrid segmentation method serves as the best.

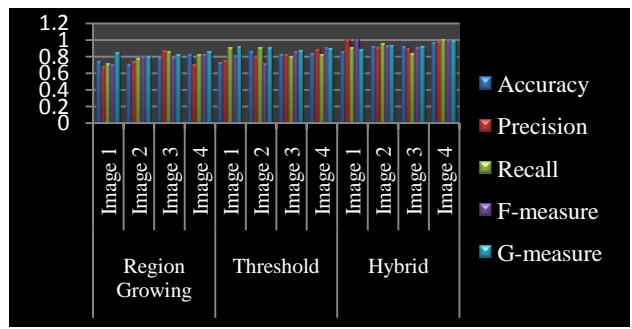

Figure 5: Evaluation graph using performance measures.

On seeing the overall performance, the new hybrid segmentation proves to be better than the other methods compared above. In figure 5, evaluation graph shows the performance of four images. So the proposed model selects & uses centre seed point and global threshold values for the process of hybrid segmentation.

## 6. CONCLUSION

The hybrid segmentation is the combination of Single seed region growing and threshold based segmentation which has been proposed to segment brain tumor images. This can be helped to improve the results of region growing segmentation. The new hybrid method is applied on several images and experiment results of performance are compared and analyzed with the ground truth image. In this work, comparative analysis of various performance metrics have been obtainable. It gives a clear decision from the experiments and performance that evaluating algorithms on an image data set leads to different ranking depending on the metrics chosen. This paper compares the performance of some segmentation algorithms such as single seed region growing, threshold segmentation and hybrid segmentation. The analysis of segmented images according to accuracy, Recall, F-measure, G-measure and precision it shows higher values. This result shows the hybrid segmentation gives better result on seeing the overall performance than two methods. The extension of the work would be the classification of tumor types with the new improved features.

## ACKNOWLEDGEMENTS

I sincerely thank to www.osirix-viewer.com, to get a brain tumor images and all my gratitude to my guide who is encouraging me to do research and my friends.

**Authors**

Short Biography

Miss. D. Anithadevi is pursuing Ph.D in Madurai Kamaraj University, Madurai. She received the M.Phil Degree in Computer Applications from Madurai Kamaraj University in the year 2014. MCA degree from Thiagarajar College of Engineering Madurai in the year 2013. She has contributed papers in International Journals and Conferences. Her research interest is in image processing and image segmentation of medical images.

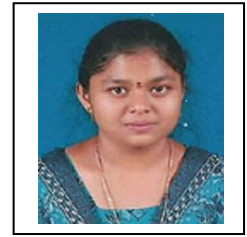

Mr. K. Perumal working as an Associate professor in Madurai Kamaraj University, Madurai. He has contributed more than 25 papers in International Journals and Conferences. He has guiding 6 scholars. His interest includes Data Mining, Image processing and medical images.

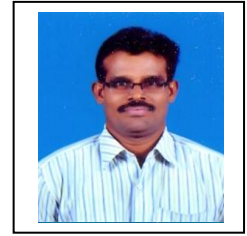